\documentclass[11pt,letterpaper]{article}
\usepackage{cogsys}
\usepackage[T1]{fontenc}
\usepackage{times}
\usepackage[pdftex]{graphicx} 

\usepackage{natbib}
\cogsysheading{X}{20XX}{1-6}{X/20XX}{X/20XX}

\ShortHeadings{R-HTN: Rebellious Online HTN Planning}
              {H.\ Munoz-Avila, D.\ W.
              \ Aha, and P.\ Rizzo}

\usepackage{balance} 

\usepackage{makecell}

\usepackage{mathtools}

\usepackage{amssymb}

\usepackage{algorithm}
\usepackage{algpseudocode}%

\usepackage{subcaption}

\begin{document} 

\newcommand{\D}{$\mathcal{D}$}

\title{R-HTN: Rebellious Online HTN Planning for Safety and Game AI}
 
\author{H\'ector Mu\~noz-Avila}{hem4@lehigh.edu}
\address{Computer Science \& Engineering,
Lehigh University;
Bethlehem, PA  18015-3084  USA}
\author{David W. Aha}{david.w.aha.civ@us.navy.mil}
\address{Navy Center for Applied Research in AI; Naval Research Laboratory; Washington DC, USA}
\author{Paola Rizzo}{p.rizzo@interagens.com}
\address{Interagens s.r.l.; Rome, Italy}
\vskip 0.2in
 
\begin{abstract}
We introduce online Hierarchical Task Network (HTN) agents whose behaviors are governed by a set of built-in directives \D. Like other agents that are capable of rebellion (i.e., {\it intelligent disobedience}), our agents will, under some conditions, not perform a user-assigned task and instead act in ways that do not meet a user's expectations. 
Our work combines three concepts: HTN planning, online planning, and the directives \D, which must be considered when performing user-assigned tasks. We investigate two agent variants: (1) a Nonadaptive agent that stops execution if it finds itself in violation of \D~ and (2) an Adaptive agent that, in the same situation, instead modifies its HTN plan to search for alternative ways to achieve its given task.
We present R-HTN (for: Rebellious-HTN), a general algorithm for online HTN planning under directives \D. We evaluate R-HTN in two task domains where the agent must not violate some directives for safety reasons or as dictated by their personality traits. We found that R-HTN agents never violate directives, and aim to achieve the user-given goals if feasible though not necessarily as the user expected.
\end{abstract}

\section{Introduction}

With the increasing availability of autonomous agents, there is also an increase in the frequency with which users direct them to perform unsafe or even dangerous behaviors. This can be intentional. For example, some recurring news items report on unmanned air vehicles (UAVs) that violate reserved air spaces or "red zones" such as airports \citep{huang2025small}. 

Sometimes, hostile actors may induce a user's errors (e.g., of omission) in directing their agents' monitoring. For instance, the actor may draw the user's focus of attention towards an unrelated event, thereby causing the user to reduce attention to their agents. More generally, unsafe use of autonomous agents may be the result of factors such as cognitive overload. This can occur when the user is performing multiple tasks, such as controlling multiple autonomous agents, and finds herself unable to monitor some actions that the agents are performing. In fact, in experiments reported by \cite{gartenberg2014situation}, users tasked with controlling multiple autonomous UAVs frequently failed to reroute them when they flew over airspace that was unexpectedly restricted (e.g., where hostile agents carrying anti-UAV weaponry were detected). In these experiments, the user needed to re-route the UAV to avoid the dangerous area. Part of the reason for the user's omission of re-routing the UAVs is that they needed to (1) focus on other tasks (e.g., positioning agents to take an aerial photograph) and (2) control the navigation of multiple UAVs simultaneously. 

Motivated by these factors, we investigate the utility of endowing online HTN planning agents with the ability to rebel. We assume that an HTN planning domain, $\Sigma$, is accompanied by a set of directives \D~(e.g., "avoid any red zone") and task correction procedures. If the agent finds itself in a state $s$ that has a discrepancy (i.e., $\delta(s) = True$), then it may attempt to repair the tasks to be executed based on its assigned task $t$, $\Sigma$ and the task correction procedures. In this situation, we distinguish between two types of rebellious HTN (R-HTN) agents:

\begin{itemize}
    \item {\bf Nonadaptive agent}: The agent stops executing task $t$ whenever it finds itself in violation of \D~ and awaits a new command from the user.
    \item {\bf Adaptive agent}: Whenever the agent finds itself in violation of \D, it will attempt to perform a corrective action (if feasible).\footnote{An example of a situation where this is infeasible is when the agent has already executed an allotted maximum number of actions to achieve $t$.}
\end{itemize}

\noindent R-HTN is motivated by the following agent design principles:

\begin{itemize}
    \item {\bf HTN planning.} A stratified planning paradigm where a task (e.g., \emph{achieve(g)}, to achieve a user-given goal $g$), is recursively decomposed into simpler tasks, until so-called primitive tasks, corresponding to actions, are generated such that when executed, these actions will fulfill $g$.
    \item {\bf Directives.} A directive is a function $\delta: S \rightarrow \{True,False\}$ such that, given a state $s \in S$, it returns {\it True} if $s$ is an unexpected state, also called a {\bf \D-discrepancy}, requiring the agent's attention (e.g., the agent finds itself inside a hazardous ("red") zone, which it is prohibited to visit).
    \item {\bf Online HTN planning.} The agent interleaves HTN planning and execution; as new states are visited, the agent detects if there is a discrepancy and may take action accordingly.
    
\end{itemize}

Our interest in rebellious agents based on HTNs is motivated by reports that HTNs are particularly suitable for many tasks, including military planning \citep{Donaldso2014}, strategic decision making (e.g., in games \citep{smith1998success,Verweij2007}), and controlling multiple agents \citep{cardoso2017multi}, including teams of UAVs \citep{musliner2010priority}.  R-HTN agents are general and can be used for any task domain provided that $\Sigma$, \D, and a task repair procedure are given.

In Section~2  we describe an example scenario that illustrates the expected behavior of an R-HTN agent. Next, we describe a taxonomy of \D-discrepancies (Section~3). Following this, we discuss online HTN planning (Section~4) and present our R-HTN agent algorithm (Section~5). Then we describe an empirical study with  two  domains and discuss the results (Section~6). We finish by discussing related work (Section~7) and providing final remarks (Section~8).

\section{Example Scenario}


Our first scenario is inspired by RESCHU's task domain \citep{boussemart2008behavioral}. RESCHU (Research Environment for Supervisory Control of Heterogeneous Unmanned Vehicles) was developed to study how cognitive overload impacts a user's decision making. A RESCHU user directs a set of UAVs to destination locations; they must also ensure that these UAVs avoid red (hazard) zones. Over time, each red zone disappears and then reappears randomly in a different location (one that does not include any agents). In our scenario, which we call O-RESCHU (for: own-RESCHU), all agents begin at the same location and start with the same number of energy points. Agents consume one energy point when moving to a contiguous cell. The user assigns a destination location to each UAV, which should avoid all red zones (trespassing red zones, and movements within a red zone, causes a UAV to lose even more energy points). 

\begin{figure}[htb]
  \centering
  \includegraphics[width=0.6\linewidth]{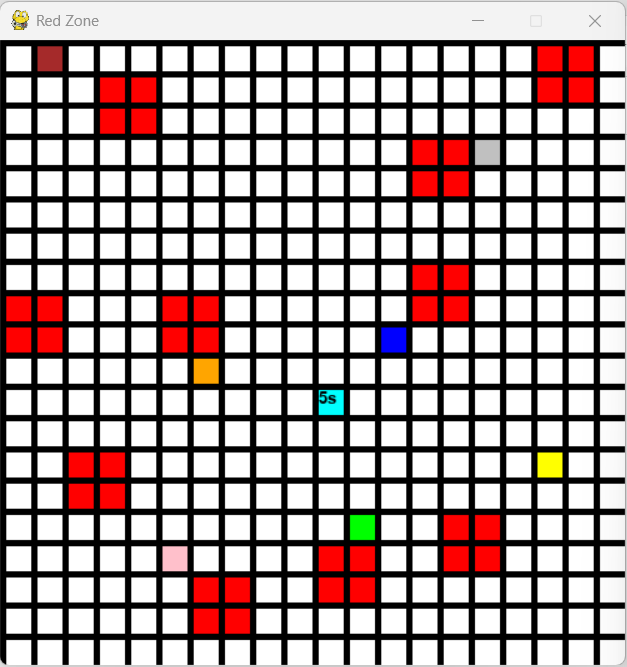}
  \caption{\normalfont A randomly generated map where all five agents start from the location labeled "5s"}
  \label{fig:RESCHUstart}
\end{figure}

The user's goal is for their agents to visit assigned destination locations while minimizing their number of expended energy points. 
The user expects each agent to follow a shortest path from its current location to the agent's user-assigned destination location, but is not notified a priori as to when and where red zones will disappear and reappear. Red zones can be thought of as meteorological conditions that cause a UAV to consume more energy than usual, hence the user should ensure that their UAVs avoid them.

\begin{figure}[htb]
  \centering
  \includegraphics[width=0.6\linewidth]{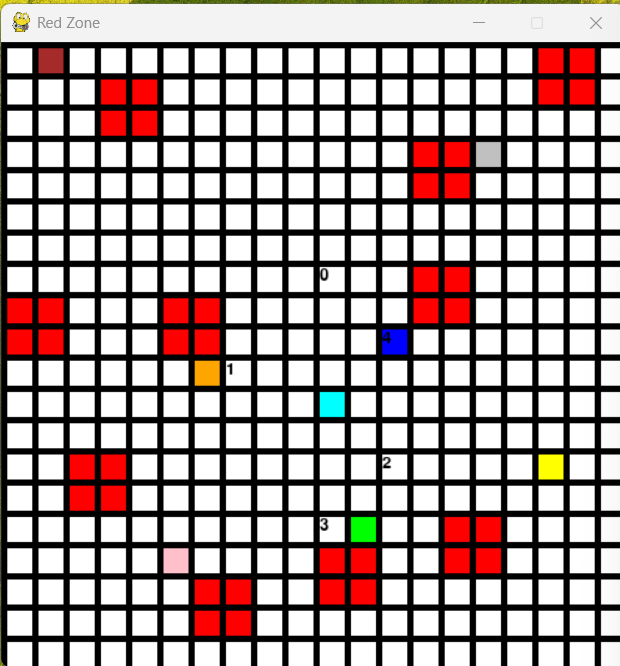}
  \caption{\normalfont Location of the five agents after four time ticks. }
  \label{fig:RESCHUafter}
\end{figure}

Figure~\ref{fig:RESCHUstart} displays a snapshot of a randomly generated configuration for O-RESCHU. The 10 (sets of) red squares are the red zones that agents should avoid (which we depict as a square of four adjacent red cells). The colored single-cell location with the label "5s" is the starting location for all five agents. The other seven colored cells are the destination locations.

Figure~\ref{fig:RESCHUafter} displays a snapshot after   four time ticks (at each tick, each agent moves at most to one adjacent cell - above, below, left or right). Agent 4 reached its (dark blue) destination. Agents 3 and 1 will reach their respective (green) and (orange) destinations in the next time tick.  Agent 0 is navigating towards the brown location (upper left).   Finally, Agent 2 is moving towards its (yellow) destination.

\section{Taxonomy}

We consider two types of agents: those that ignore \D~and those that heed \D~during their reasoning process. Agents that attempt to perform the user's assigned tasks without any other consideration are called {\bf Compliant} agents \citep{coman2014motivation}. In our work, Compliant agents ignore \D, and therefore generate and execute a plan based solely on the HTN domain $\Sigma$. Among those that heed \D, we distinguish between ways these agents respond to \D-discrepancies (see Table \ref{table:taxonomy}). 

The first row is for agents that test whether a discrepancy exists in the current state $s$  (we call this a \D-{\it discrepancy}), such as when an agent finds itself in a red zone cell. Let $s$ be a collection of grounded predicates. For example, $s = \{ at(5,(10,11)), red(7, (10,11), 2) \}$ indicates that agent $5$ is located in cell $(10,11)$. It also indicates that the red zone number $7$'s is at $(10,11)$ and the redzone is of size $2 \times 2$.  Let \D ~be  a collection of 10 directives, $\delta_0,\dots,\delta_9$, one per red zone. If $\delta_i(s) = True$, it means at least one of the agents finds itself inside red zone $i$ in state $s$. Therefore, in this situation, $\delta_7(s) = True$ because agent $5$ is inside redzone $7$.

The second row in Table~\ref{table:taxonomy} is for agents that project whether a \D-discrepancy will occur in a  future state. We call this a {\it projected \D-discrepancy}. Let $s$ again be a set of grounded predicates. Let $\pi_n = (a_1 a_2 \dots a_n)$ be the next $n$ actions to be executed by the agent. Let $\pi_n(s)$ be the state resulting from applying the next $n$ actions, starting in state $s$:

\begin{equation}
    \pi_n(s) =
    \begin{cases*}
      a_1(s)  & if $n = 1$\\
      a_n(s') & if $n > 1$ and $s' = \pi_{n-1}(s)$
    \end{cases*}
\end{equation}

\noindent In this case, a projected \D-discrepancy occurs if, for $n \geq 1$, $\delta(\pi_n(s)) = True$.

The third case, {\it Adaptive \D-discrepancy}, refers to an agent that generates a plan $\pi'$, a repair of plan $\pi$, such that there is no projected \D-discrepancy during the execution of its next $n$ actions and $\pi'$ achieves a user-given goal $g$. This is formalized in the last column of Row 3 in Table~\ref{table:taxonomy}.

\begin{table}

\begin{center}
 \caption{Taxonomy of \D-discrepancies}
\label{table:taxonomy}
\begin{tabular}{||c | c | c||} 
 \hline
 Type & Example & Formal description \\ [0.5ex] 
 \hline\hline
 Immediate $\mathcal{D}$-discrepancy  & in red zone &  $\delta(s) = True$ \\ 
 \hline
 Projected $\mathcal{D}$-discrepancy & \makecell{will reach\\ red zone} &  
 $\delta(\pi_n(s)) = True$ \\
 \hline
 Adaptive $\mathcal{D}$-discrepancy & \makecell{turn to avoid\\ red zone}  &  \makecell{$\delta(\pi'_n(s)) = False$\\ and $\pi'(s) \vDash g$} \\ [1ex]
 \hline
\end{tabular}
\end{center}
\end{table}

\section{HTN Planning}

Hierarchical Task Network planning is a planning paradigm in which complex tasks are decomposed into simpler tasks until a sequence of so-called primitive tasks is generated \citep{georgievski2015htn}. Each primitive task is achieved by an {\it action}; hence, the output of HTN planning is a sequence of actions $\pi$. We use $T$ to denote the collection of all primitive and nonprimitive tasks. An HTN planning model, $\Sigma$, is a collection of {\it actions} and {\it methods}, which we define next. 

An action $a$ defines the usual transition function $a\!: S \rightarrow S \cup \emptyset$, where $S$ is the set of all states. If $a(s) = \emptyset$, then $a$ is not applicable in $s \in S$. Otherwise, $a(s) \in S$ is the resulting state after executing action $a$ in $s$. 

A method $m$ is used to decompose a nonprimitive task $t$. A method $m\!: S \times T \rightarrow \tilde{T} \cup \emptyset$, where $\tilde{T}$ is the set of all possible sequences of tasks in $T$.  If $m(s,t) = \emptyset$, then $m$ is not applicable in $(s,t)$. Otherwise, $m(s,t) \in \tilde{T}$ is a sequence of tasks. In that case, we say that $t$ is decomposed into $m(s,t)$.

To exemplify a method, {\it navigate-distant}, let $s$ be the state displayed in Figure \ref{fig:RESCHUafter}. {\it navigate-distant}  is used to accomplish {\it reach(agent,location)} whenever a destination {\it location} is at least 2 cells distant from {\it agent}. In this state, Agent $0$ is currently in a location $(x,y)$. This agent has been assigned to reach the {\it brown} destination location (near the upper left corner). For this particular state, the {\it navigate-distant} method transforms the nonprimitive task {\it reach(0,brown)} into two subtasks. The first is a primitive task {\it up(0)}, which executes the action moving Agent $0$ to $(x,y-1)$, and the second task is a recursive call to the nonprimitive task {\it reach(0,brown)}. Succinctly, {\it navigate-distant(reach(0,brown)) = (up(0), reach(0,brown))}. 

To exemplify another method, {\it navigate-close}, consider Agent 3, which has been assigned the green location as its goal. That is, Agent 3 must achieve the nonprimitive task {\it reach(3,green)}. In this case, the {\it navigate-close} method will decompose the task into a single primitive task {\it right(3)}, which (when executed) moves Agent 3 to its goal.  Succinctly, {\it navigate-close(s,reach(3,green)) = (right(3))}. 

 In summary, we described two methods, one for situations when the agent's destination is at least two cells away and upwards from the agent location and another one for when the agent's destination is adjacent to its current location. But not the other way around.  That is, the {\it navigate-distant} method is not applicable for states where the agent is adjacent to the goal location. Similarly, {\it navigate-close} is not applicable in states where the agent is not adjacent to the goal location. Therefore,

 \vspace{6pt}

{\it navigate-distant(s,reach(3,green)) = $\emptyset$} 

{\it navigate-close(s,reach(0,brown)) = $\emptyset$} 

 \vspace{6pt}

\section{Rebellious Online HTN Planning}

Algorithm \ref{alg:ChatHTN} displays the pseudocode for R-HTN. This pseudocode is based on HTN planning as in the SHOP system \citep{nau1999shop}. The non-underlined parts are the standard SHOP pseudocode and the underlined  parts are our additions.

\begin{algorithm}[htb]
\caption{The R-HTN algorithm}
\label{alg:ChatHTN}
\begin{algorithmic}[1]
    \Procedure{R-HTN}{$s,\tilde{t}$}
\State \textbf{return} RSeekPlan($s,\tilde{t},())$ \Comment{() is the empty plan; a plan with no actions}
\EndProcedure
\State
\Procedure{RSeekPlan}{$s,\tilde{t},\pi$} \Comment{$\pi$ is the plan generated so far}
\State \textbf{if} $\tilde{t} =  ()$ \textbf{then return} $()$ \Comment{returns the empty plan}
\State let $\tilde{t} = (t_0, t_1,...,t_n)$
\State \textbf{if} $t_0$ is primitive \textbf{then}
\State \ \ \ \  let $a_0$ be the action associated with $t_0$
\State \ \ \ \  \textbf{if} $a_0(s) = \emptyset$  \textbf{then return} $\emptyset$
\State \ \ \ \  \underline{$\tilde{t}' \leftarrow$ RepairTasksIfNeeded($\tilde{t}$, $s$, $a_0$)}
\State \ \ \ \  \underline{\textbf{if} $\tilde{t'} \neq \tilde{t} $ \textbf{then } } \Comment{D-discrepancy occurs}
\State \ \ \ \ \ \ \ \ \ \underline{\textbf{return}  
 \textsc{RSeekPlan}$(s,\tilde{t}',\pi)$}\Comment{recursive call}
\State \ \ \ \ \underline{$s'\ \leftarrow$ \textbf{execute} $a_0$ on $s$}

\State \ \ \ \    \textbf{return}  
 \textsc{RSeekPlan}$(s',(t_1,...,t_n),\pi \cdot (a_0)$)\Comment{recursive call}
\State \textbf{if} $t_0$ is compound \textbf{then}
\State \ \ \ \  \textbf{for} $m_0 \in M$ \textbf{do} \Comment{$M$ is the list of all methods}
\State \ \ \ \  \ \ \ \ \textbf{if} $m_0(s,t_0) \neq \emptyset$ \textbf{then}
\State \ \ \ \ \ \ \ \ \ \ \ \ \ \ let $\pi'$ =
 \textsc{RSeekPlan}$(s,m_0(s,t_0) $  $\cdot (t_1,...,t_n), \pi)$ 
\State \ \ \ \ \ \ \ \ \ \ \ \ \ \ \textbf{if} $\pi' \neq \emptyset$ \textbf{then}
\State \ \ \ \ \ \ \ \ \ \ \ \ \ \ \ \ \ \textbf{return} $\pi'$

\State \textbf{return} $\emptyset$
\EndProcedure
\State
\Procedure{\underline{repairTasksIfNeeded}}{\underline{$\tilde{t}$, $s$, $a_0$}}
\State \underline{\textbf{for} $\delta \in $ \D ~ {\bf do}}
\State \ \ \ \ \underline{if $\delta(s)$ then} \Comment{$\delta$ is violated in state $s$}
\State \ \ \ \ \ \ \ \ \underline{\textbf{return} repairTaskListState($\delta$,s,$\tilde{t}$)} \Comment{an updated list is returned}
\State \underline{\textbf{for} $\delta \in $ \D ~ {\bf do}} \Comment{only check if no violations in current state $s$}
\State \ \ \ \ \underline{if $\delta(a_0(s))$ then} \Comment{$\delta$ is violated in state $a(s)$}
\State \ \ \ \ \ \ \ \ \underline{\textbf{return} repairTaskListEffect($\delta$,s,$\tilde{t}$,$a_0$)} \Comment{an updated list is returned}
\State \underline{\textbf{return} $\tilde{t}$} \Comment{no directives violated; return task list unchanged}

\EndProcedure

\end{algorithmic}
\end{algorithm}

The R-HTN planning procedure R-HTN$(s,\tilde{t})$ receives as input a state $s$ and a task list $\tilde{t} \in \Tilde{T}$ (Line 1). It calls RSeekPlan with the same parameters as R-HTN plus an empty list. The empty list represents the empty plan (i.e., a plan with no actions). RSeekPlan recursively generates the solution plan $\pi$ by decomposing $\tilde{t}$ as fgoldollows. If the task list $\tilde{t}$ is empty (i.e., $()$), it returns the empty plan (Line 6). Otherwise, $\tilde{t} = (t_0, t_1, \dots, t_n)$ is a nonempty list of tasks (Line 7). Then there are two cases:

\begin{itemize}
    \item (Case 1) if $t_0$ is primitive and its associated action $a_0$ is applicable in $s$ (i.e., $a_0(s) \neq \emptyset$; Line 10), then check if repairs are needed and return a task list $\tilde{t'}$ (line 11; we will expand on this later).  If $\tilde{t}$ differs from $\tilde{t'}$, then $\tilde{t'}$ repairs $\tilde{t}$ and planning  proceeds recursively with $\tilde{t'}$ (Lines 12 and 13) with $\pi$ and $s$ unchanged. Otherwise, $a_0$ is executed, resulting in a new state $s'$ (Line 14). Then planning  proceeds recursively with the new state $s'$, the remaining task list $(t_1, \dots, t_n)$, and the plan $\pi$ augmented with $a_0$ (Line 15).
    
     \item (Case 2) if $t_0$ is non-primitive and there is a method $m$ applicable to $s$ and $t_0$ (i.e., $m(s,t_0) \neq \emptyset$; Lines 16-18), then continue planning recursively with the same state $s$ and plan $\pi$, and the augmented task list  $m(s,t_0) \bullet (t_1,...,t_n)$, where $\bullet$ denotes a concatenation of task lists.\footnote{i.e., $(t_0,...,t_n) \bullet (t'_0,...,t'_m) = (t_0,...,t_n, t'_0,...,t'_m)$} If the recursive call returns a non empty plan $\pi'$, this is returned (Line 21).
     
\end{itemize}

Line 22 is the catch-all case where either $t_0$ is primitive but Case 1 yields no solution or $t_0$ is compound and Case 2 yields no solution. In either situation the procedure returns $\emptyset$, denoting that no such plan exists.

\paragraph{\bf Online HTN Planning}

We focus on dynamic environments, where the world state changes as a result of the agent's own actions and also as a result of factors independent of the agent's own actions. For instance, it may begin to rain in a particular area (and therefore it becomes a red zone) or it stops raining in another area (and therefore it is no longer a red zone). To respond appropriately to these situations, R-HTN  uses online planning in Line 14. It executes the chosen action $a_0$ and returns the state $s'$ observed after executing $a_0$ in the environment (we describe the simulators for the two domains in Section~6).


\paragraph{The task repair procedure.} 

Lines 25-33 detail the RepairTasksIfNeeded procedure. It receives the current task list $\tilde{t}$, the current state $s$ and the action to be executed $a_0$ (Line 25). It first checks if a \D-discrepancy exists in current state $s$ (Lines 26 and 27). If so it calls the domain-specific procedure  repairTaskListState, which returns an updated task list (Line 28). If there are no \D-discrepancies in the current state, it checks if any \D-discrepancies to the projected state $\delta(a_0(s))$ exist (Lines 29 and 30). If so, it calls the domain-specific procedure repairTaskListEffect, which returns an updated task list. The call to $a_0(s)$ does not change the state of the world. This is just standard planning to compute the projected next state. In R-HTN the only time an action is executed in the environment is in Line 14. In Section 6, we will describe our task repair procedures for the two task domains. 


\section{Empirical Evaluation}

We conducted an evaluation in which we tested three online HTN planning agents: Compliant, Nonadaptive and Adaptive, as described in earlier sections. In these experiments, a simulated user randomly selects the destination locations for each agent to visit. 

Our hypotheses are as follows:

\begin{itemize}
    \item Adaptive will achieve more goals compared to Nonadaptive and Compliant.
    \item Adaptive and Nonadaptive will incur in no state violations.
\end{itemize}

We have two domains:

\begin{itemize}
    \item {\bf O-RESCHU}. The domain we have discussed so far in the paper. The simulated user never assigns two or more agents to the same location.
    \item {\bf MONSTER}. To make NPCs more realistic, game designers created the idea of NPC alignment \citep{gygax1978advanced}. NPC alignment refers to personality traits such as "good" or "evil" with the idea that if the player orders an NPC to perform an action against its alignment (e.g., order an NPC to steal gold but the NPC has a "good" alignment), the NPC may refuse the order. 
    MONSTER is a straighforward modification of our O-RESCHU simulation: the red zones are re-interpreted as monsters and colored locations are reinterpreted as gold locations,  each having 5 gold coins. The (simulated) user sends the NPC agent to these locations to collect gold. Following the conventions of many games, an NPC agent collects gold by navigating to a cell with gold (i.e., there is no explicit action to collect the gold). Once collected, the location will no longer have any gold. NPCs and monsters each begin with 10 health points (hps). When a monster and an NPC are in the same cell, they "fight to the death": namely, we repeatedly roll a die such that with 50\% probability the monster loses 1 hp whereas with 50\% probability the NPC loses 1 hp. Once either reaches 0 hp, it is eliminated from the game. If the NPC reaches 0 health, the simulation terminates as there is only one NPC acting per game. If the monster reaches 0 health, the monster disappears from the game. The NPC continues with whatever health it has left after the fight. This means that, on average, an NPC will survive fewer than one fight with a monster, and if it does survive, it will typically have very little health left, making it unlikely to win a second fight. Monsters do not move but they can respawn in a different location although never in an NPC's cell. They can respawn on a gold location. 
\end{itemize}

\subsection{\bf Task Repair}

\paragraph{O-RESCHU.}
Reconsidering the scenario in Figure \ref{fig:RESCHUafter}, suppose Agent 3 (which is adjacent to the green destination location) is instead assigned to navigate to the pink location (i.e., its revised goal). Again, suppose that the only directive in \D~ is for the agent to avoid any red zone, and that its current task list is: {\it (down(3), reach(3,pink))}.

In this situation, there will be a projected \D-discrepancy with $n=1$ (i.e., applying $\pi_1 = (down(3))$ results in a \D-discrepancy). To address this situation, our agent will repair the task list by considering the current state $s$ and goal $g$. We denote by $A(s)$ the alternative actions  applicable in $s$ such that applying them does not yield a \D-discrepancy. In this case, $A(s) = \{ up(3), left(3), right(3) \}$. The alternative selected action $a$ is one satisfying:

\begin{equation} 
a = min_{a \in A(s)} dist(a(s),g),
\end{equation}

\noindent where $dist$ is the distance function (in this case we use Manhattan distance) and {\it min} is the minimum.

In this scenario, $a = left(3)$ is selected because it moves Agent 3 to the closest location to the pink destination, in comparison to the two alternatives. Hence, Agent $3$ will modify the task list by replacing $down(3)$ with $left(3)$. This yields a modified task list: {\it (left(3), reach(3,pink))}. Our agent will then continue executing the online HTN planning process, moving Agent 3 to the left once and then recursively performing online planning with {\it reach(3,pink)}.

Because this agent is planning online, it will appropriately respond to new contingencies such as when a red zone spawns on its path. Analogously, if the red zone that caused a task repair disappears, then the agent will consider the newly viable path to its goal.

In our scenarios, the agent is never "boxed" in red zones. Red zones are always squares and placed randomly on the map but such that there is a gap between them. This means that the agent can always navigate between red zones to reach any location. In other task domains where the agent {\it can} be boxed in, no matter which alternative it takes, then $A(s)$ would be defined as the set of all first actions in feasible plans for reaching $g$. If $A(s)$ is empty, then our agent will cease its execution.

\paragraph{MONSTER.} It has only one directive, $\delta_{monster}(s)$, which is true if it finds itself in the same cell as a monster. Since the agent checks if the next action will result in a \D-discrepancy (Line 30 of the algorithm) and monsters never spawn in the same cell as an NPC, it means that the our Adaptive NPC agent will know in advance if it will encounter a monster along its path to collect gold. 
Our Adaptive NPC agent exhibits a cowardly trait because it avoids fights with monsters.
That is, it will modify its task list in the exact same manner as task repair in O-RESCHU (i.e., it will move around the monster). If the assigned gold location has a monster, the agent will abandon its user-assigned task, resulting in rebellion but consistent with its cowardly personality trait.

\subsection{\bf Task Domain Map and Settings.}

\paragraph{O-RESCHU.}
Our map is a 2D grid of 20 $\times$ 20 cells. It contains 10 red zones of size 2 $\times$ 2, 7 (colored) destination locations (i.e., cells), and 1 initial start location for all 5 agents. These settings are illustrated in Figures \ref{fig:RESCHUstart} and \ref{fig:RESCHUafter}. The destination locations and start location spawn randomly at the beginning of each episode and remain fixed throughout the episode.
Red zones respawn with a fixed probability. In our experiments, we use the probabilities: 0 (i.e., red zones remain fixed throughout the episode), 5, 10, 15, 20, 25, 30, 35, 40, 45, and 50 (i.e., red zones will relocate 50\% of the time, therefore once every two time ticks on average). Red zones never spawn on top of agents, although they may spawn on top of destination locations. Agents cannot move diagonally. Agents may stay in their current location without moving.

\paragraph{MONSTER.}
 The map has the same dimensions and number of red zones (which here are monsters) as O-RESCHU, and colored locations (which contain gold). Monsters spawn with the same probability setting as in O-RESCHU.  The only difference is that there is a single agent acting in the grid. 

For both domains, for each red zone (or monster) respawn probability and each of the 3 agents, we ran 100 episodes and calculated the average of the dependent measures, which we describe next.

\subsection{\bf Dependent Measures.}

 \paragraph{Common for both domains.} The first is a binary measure indicating whether the agent achieved its goal (i.e., it reached its assigned location). The second is a counter of 
 how many \D-discrepancies the agent incurred. 
 
 \paragraph{O-RESCHU.} For this task domain, we also measure the number of penalty points an agent incurs. Each time an agent moves one step within non-red cells it incurs a penalty of one point. Each time it instead moves one step into, out from, or between red cells it incurs a penalty of 20 points. At the start of an episode, each agent is allotted 38 points, representing the maximum number of steps needed to navigate between any two locations in the map, if following a shortest path. An agent ceases movement in an episode whenever it has 0 remaining points.

 \paragraph{MONSTER.} In each episode, the simulated user will assign a gold location, followed by a second location after the agent reaches the first location or abandons trying to reach it. For each game, we compute two additional dependent measures.  The first is the total gold collected, which is a number between 0 and 35 (e.g., if the gold locations are clustered together along a shortest path to its destination such that the  agent visits them all). The second measure is the number of agent deaths (either 0 or 1).  

 \subsection{Results O-RESCHU}

 \begin{figure}[htb]
  \centering
  \begin{subfigure}[t]{0.48\linewidth}
    \centering
    \includegraphics[width=\linewidth]{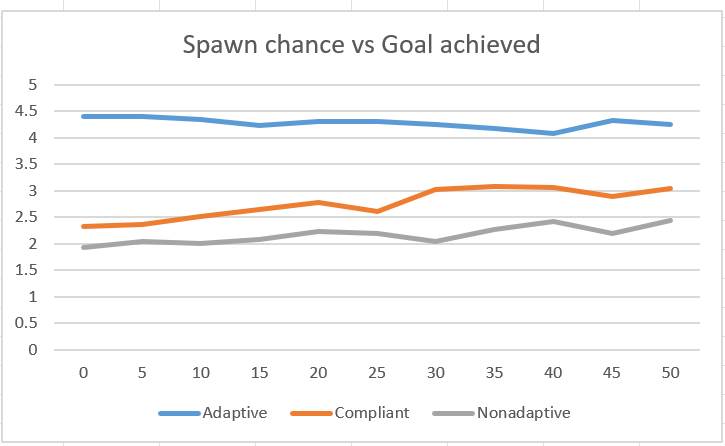}
    \caption{\normalfont The horizontal axis denotes the probability that the red zones spawn (with each tick). The vertical axis denotes the number of goals achieved, averaged over 100 episodes per agent type.}
    \label{fig:GoalsAchieved}
  \end{subfigure}%
  \hfill
  \begin{subfigure}[t]{0.48\linewidth}
    \centering
    \includegraphics[width=\linewidth]{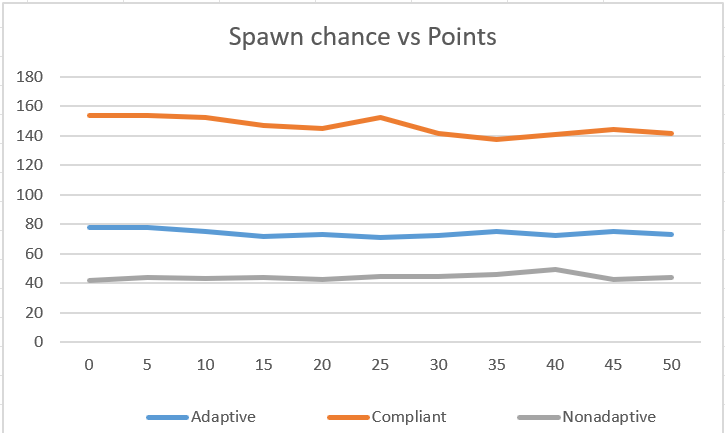}
    \caption{\normalfont The horizontal axis denotes the probability that the red zones spawn (with each tick). The vertical axis denotes the number of penalty points incurred, averaged over 100 episodes per agent type.}
    \label{fig:Points}
  \end{subfigure}
  \caption{Performance measures as a function of red zone spawn probability.}
  \label{fig:sidebyside}
\end{figure}

\begin{figure}[htb]
  \centering
  \includegraphics[width=0.5\linewidth]{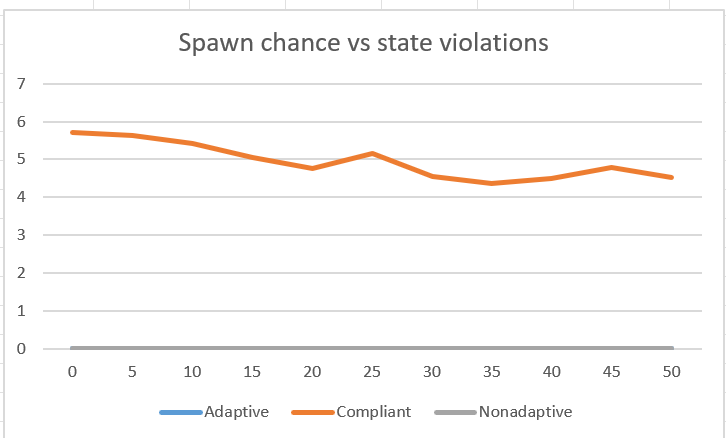}
  \caption{\normalfont The horizontal axis denotes the probability that the red zones spawn (with each tick). The vertical axis denotes the  number of \D-discrepancies incurred, averaged over 100 episodes per agent type.}
  \label{fig:Violations}
\end{figure}

O-RESCHU's results are displayed in Figures \ref{fig:sidebyside}\subref{fig:GoalsAchieved}, \ref{fig:sidebyside}\subref{fig:Points}, and \ref{fig:Violations}. In each the x-axis denotes the probabilities that red zones will be respawned (from 0\% to 50\%) in each time tick. To account for randomness each data point is the average over 100 runs. 

\paragraph{\bf Experimental Results - Goals Achieved:}
As displayed in Figure \ref{fig:GoalsAchieved}, the Adaptive agents achieve the highest average number of goals per episode, which varies between 4 and 4.5. The maximum number of goals is 5 (i.e., one per agent). There are two reasons why, on average, these never reach 5 for the Adaptive agents. First, the red zone will sometimes spawn on top of an assigned destination location, preventing such agents from reaching them (i.e., to not violate \D). Second, an agent may run out of points while navigating around red zones. Compliant agents achieve an average of 2 to 3 goals per episode. This number increases with red zone spawn frequency because red zones sometimes move out of an agent's path to a goal location. Also, Compliant agents can move to a destination location that is inside a red zone. Finally, Nonadaptive agents achieve an average of 2-2.5 goals per episode. Nonadaptive agents achieve fewer goals than Compliant agents because the former always abandons a goal if moving will result in a violation whereas the latter will still move and sometimes still   reach the goal. This experiment confirms our hypothesis that Adaptive  achieves more goals than Nonadaptive and Compliant.

\paragraph{\bf Experimental Results - Penalty Points Incurred:}

Figure \ref{fig:Points} displays the average number of penalty points incurred. Compliant agents incur the largest average number of penalty points (140-160), almost doubling that of the Adaptive agents. These averages trend  slightly downwards for Compliant agents because red zones sometimes move out of their planned path. Nonadaptive agents incur fewer penalty points because they more frequently abandon their goals, as shown in Figure \ref{fig:GoalsAchieved}.

\paragraph{\bf Experimental Results - State Violations Incurred:}
Figure \ref{fig:Violations} displays the average number of \D-discrepancies incurred per agent type. Adaptive and Nonadaptive agents never incur a \D-discrepancy. In contrast, Compliant agents incur an average of 4-6 state violations per episode. Again, this decreases as the probability of respawning increases because paths clear of red zones become available more frequently as red zones spawn (and de-spawn) with more frequency. 
This experiment confirms our hypothesis that Adaptive and Nonadaptive incur in no state violations.

\medskip

Overall, Adaptive agents achieve (on average) more goals compared to the other agents, while incurring no state violations and incurring approximately only half of the number of penalty points incurred by Compliant agents. The Nonadaptive agents incur the fewest penalty points but also achieve fewer goals compared to the other agents.

\subsection{Empirical Results - Monster}

 \begin{figure}[htb]
  \centering
  \begin{subfigure}[t]{0.48\linewidth}
    \centering
    \includegraphics[width=\linewidth]{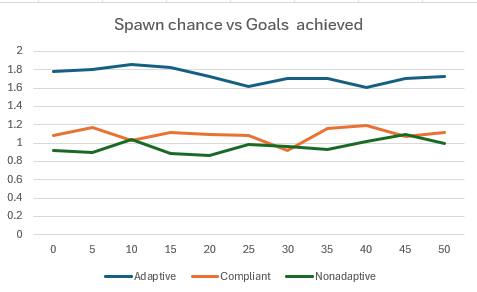}
    \caption{\normalfont The horizontal axis denotes the probability that the monsters spawn (with each tick). The vertical axis denotes the number of goals achieved, averaged over 100 episodes per agent type.}
    \label{fig:MONSTERGoalsAchieved}
  \end{subfigure}%
  \hfill
  \begin{subfigure}[t]{0.48\linewidth}
    \centering
    \includegraphics[width=\linewidth]{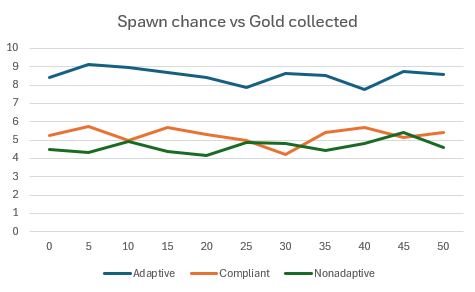}
    \caption{\normalfont The horizontal axis denotes the probability that the monsters spawn (with each tick). The vertical axis denotes the total gold collected, averaged over 100 episodes per agent type.}
    \label{fig:GoldCollected}
  \end{subfigure}
  \caption{Performance measures as a function of monster spawn probability.}
  \label{fig:MONSTER1}
\end{figure}

\begin{figure}[htb]
  \centering
  \begin{subfigure}[t]{0.48\linewidth}
    \centering
    \includegraphics[width=\linewidth]{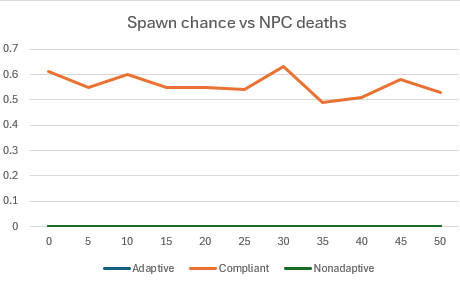}
    \caption{\normalfont The horizontal axis denotes the probability that the monsters spawn (with each tick). The vertical axis denotes the number of agents deaths, averaged over 100 episodes per agent type.}
    \label{fig:AgentsDeaths}
  \end{subfigure}%
  \hfill
  \begin{subfigure}[t]{0.48\linewidth}
    \centering
    \includegraphics[width=\linewidth]{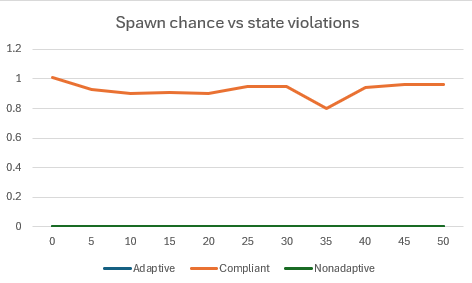}
    \caption{\normalfont The horizontal axis denotes the probability that the monsters spawn (with each tick). The vertical axis denotes the number of state violations, averaged over 100 episodes per agent type.}
    \label{fig:MonstersKilled}
  \end{subfigure}
  \caption{Performance measures as a function of monster spawn probability.}
  \label{fig:MONSTER2}
\end{figure}

\paragraph{\bf Experimental Results:}
Figure  \ref{fig:MONSTER1}\subref{fig:MONSTERGoalsAchieved} displays the average number of goals achieved by each agent. The adaptive agent (i.e., Coward) achieves the highest average; the Compliant achieves slightly more goals than the Nonadaptive agent. 
This experiment confirms our hypothesis that Adaptive  achieves more goals than Nonadaptive and Compliant. 
Similarly, as shown in Figure  \ref{fig:MONSTER1}\subref{fig:GoldCollected}, the adaptive agent collects more gold than the Compliant and Nonadaptive agents, which have similar performance on this measure.
The reason for this difference between Nonadaptive and Adaptive, for both measures,  goals achieved and gold collected, is that Nonadaptive agents stop when they encounter any monster. In contrast Adaptive agents will seek to go around any monster unless it is
located on top of the gold location. The difference between Compliant and Adaptive is due to  Compliant dying more frequently, as it confronts any monster on its path to its assigned gold location, as illustrated in Figure  \ref{fig:MONSTER2}\subref{fig:AgentsDeaths}, which shows that Compliant dies in more than half of the episodes. This is also illustrated in Figure  \ref{fig:MONSTER2}\subref{fig:MonstersKilled}, which shows that Compliant has close to one \D-discrepancy per episode. In contrast, Adaptive and Nonadaptive never incur in \D-discrepancies, thereby confirming our experimental hypothesis that these two agents do not result in state violations.

Overall, we observe from the  measures that the agents' performance is consistent with their traits. The adaptive (Coward) agent never dies and collects the most gold (though it does not always achieve its assigned goal). The nonadaptive agent also never dies, but since encountering any monster will result in the agent forgoing achieving its goal, it collects less gold. Finally, the Compliant agent will "mindlessly" pursue its assigned gold, thereby dying more frequently, and as a result collects less gold than the adaptive agent.

\section{Related Work}

Our model of rebellion can be connected to theories of human cognition. In human factors research, \emph{intelligent disobedience} \citep{chaleff2015intelligent} describes how operators or assistants are trained to override commands when they conflict with safety rules. A well-known example is guide dogs, which are trained to refuse an owner’s command if obeying would place the owner in danger. For instance, the dog will stop and refuse to move forward at a busy intersection. Similarly, in the military domain, a disciplined soldier should refuse orders if complying would result in crimes against humanity. In our work, we endow agents with directives $D$ such that when a user’s commands would result in a violation, the agents adapt to avoid creating the violation while still aiming to fulfill the user’s intent. This is analogous to the dog waiting until it is safe to cross the street. Chaleff (\citeyear{chaleff2015intelligent}) refers to this behavior as “pause and assess”, highlighting the principle that intelligent disobedience involves deliberation rather than blind obedience.

 Psychological models of cognitive control (e.g., conflict monitoring in executive function) explain how humans detect contradictions between goals and constraints and engage in alternative planning \citep{greene2001fmri}. An illustrative case is the moral dilemma faced by an autonomous vehicle with one passenger on board; it must choose between deviating from its route, potentially harming its passenger, or continuing and potentially harming multiple pedestrians at an intersection. In our context, this corresponds to situations where two or more directives in $D$ are simultaneously triggered, requiring the agent to prioritize among competing norms or ethical constraints. In the current version of our work, the agent resolves such conflicts by selecting one directive arbitrarily (Lines 28 and 31). In future work, we will explore mechanisms for resolving directive conflicts in a more principled manner, drawing inspiration from models of human moral reasoning.

The term {\it rebel agents} was coined by Coman and Mu\~{n}oz-Avila (\citeyear{coman2014motivation}). That work was motivated by digital games and the need for NPCs that behave in more sensible ways when receiving a player's commands. It used the idea of a drama manager \citep{yu:aiide2013} that assesses whether the agent's motivations conflict with the user's commands. For instance, the user might command the agent to get an item by stealing it but this contradicts with the agent's "lawful good" alignment (\cite{gygax1978advanced}; page 23). Coman and Mu\~{n}oz-Avila (\citeyear{coman2014motivation}) also established the basis for the rebel agents where the agent checks the results of planning versus the agent's motivations.  In this paper, we instead formalized the ideas of \D-discrepancies in the context of online HTN planning. In prior work,  researchers defined rebel agents in the context of goal-driven autonomy \citep{molineaux2010goal} models of goal reasoning. Hence, when a \D-discrepancy occurs, those models generate a new goal. In contrast, our Adaptive agents will still attempt to achieve a goal by adapting the plan in a manner that does not contradict \D~(i.e., they will attempt to replan rather than abandoning the goal). More recently, there has been increased interest in the subject of rebel agents. For a broader discussion see \citep{coman2018ai}. 

Our work is also related to norms in multi-agent systems (MAS) \citep{sergot2007action}. Norms are  principles that an agent is expected to abide by when executing actions in an environment. A large body of work exists on the study of norms for MAS \citep{savarimuthu2011norm}. In our work, we do not prescribe how \D~can be generated. Hence, it is conceivable that \D~captures norms for a given domain. The only requirement that we have is that \D~can be tested to determine if it is violated in a state. In this way, agents can conduct tests for the current or projected states.

Our work is also related to BDI (belief-desire-intention) models \citep{georgeff1999belief}. These models are designed to monitor and ensure that a plan's execution fulfills its intent and annotate alternatives when these intents are not fulfilled. In our work, we use HTNs for plan generation; the resulting task hierarchies and \D~perform a similar role during plan generation. For example, the action {\it down(3)} can be explained because the agent is attempting to achieve the task {\it reach(3,pink)}. If there is a \D-discrepancy, then HTN replanning can be used to revise the plan. 

Our online HTN planning approach is consistent with the actor's view of planning paradigm \citep{ghallab2014actors}. This paradigm advocates using HTN planning to interleave planning and execution precisely for the kinds of environments we use in this paper where changing state dynamics may require an agent to perform replanning during plan execution. As an example, in \cite{yuan2022task}, HTN plan generation is interleaved with plan execution resulting in task modifications when actions become inapplicable as a result of the dynamic changes in the environment. But that work does not consider directives, so the  task modifications are due to changing conditions in the environment as opposed to  a violation of \D. Another difference is that that work does not perform lookahead to consider violations in future states. 

\section{Conclusions}

We presented rebellious online HTN planning agents, which are agents capable of reasoning  with directives \D. Directives enable these agents to assess if state discrepancies will occur, and replan as needed. We described how HTN planning agents can use \D~to prevent such discrepancies from arising for safety domains and how can they use \D~to endow NPCs with personality traits. From experiments we conducted using the O-RESCHU task domain, we confirmed that these agents never violate \D. Furthermore, we explored two variants: Nonadaptive and Adaptive, where the latter (as needed) dynamically modifies its task list to generate alternative plans that do not violate \D. We found that Adaptive agents achieve most of their assigned goals with lower costs (i.e., number of penalty points) compared to Compliant agents. The Nonadaptive agents incur the least cost but also achieve the fewest number of goals. For the MONSTER domain, we observed similar results, with the adaptive agent collecting more gold without ever dying, while the Nonadaptive agent collects less gold, although also never dying, and the compliant agent dying frequently and also collecting less gold. 

Potential future work includes the study of agents that look ahead before executing a larger number of actions. This is needed in domains where an agent may find itself unable to backtrack. For instance, in a variant of the RESCHU domain an agent may not be permitted to retrace its steps and red zones may spawn in overlapping areas. In this case, an agent will need to generate $n$ steps ahead, $\pi_n$ (with $n \geq 2$), to check for possible \D-discrepancies without executing the actions. Another potential future work direction is to consider \D~with numeric fluents (e.g., gasoline consumption) and agents  that can explicitly reason about their associated \D-discrepancies. We will also study nondeterminism in this context, where actions have more than one possible outcome. For example, due to wind conditions, when executing the action "up" from a location in cell $(x,y)$, an agent may find itself in location $(x,y-1)$ or $(x+1,y-1)$. We believe that R-HTN would work as-is if a potential outcome of the action incurs a \D-violation (e.g., if cells $(x,y)$ or $(x+1,y-1)$ are in a red zone, then the agent will consider alternative actions). However, if the agent needs to reason with probabilistic outcomes (e.g., where a 95\% chance of not violating \D~is considered acceptable), then further research will be needed.  In this case, the agent has to reason with the probability of reaching a state where violations occur. Additional future work includes an analysis of the computational gains of online HTN planning versus linear replanning, through comparisons involving different solvers in terms of replanning frequency and required computational time.

Another measure worth investigating is the amount of deliberation required to generate rebellious behavior with different types of constraints. For instance, the directive to "avoid red zones" (due to loss of energy points) can be compiled away as a safety constraint that does not require exceptional deliberation beyond constraint satisfaction. However, a rebellious behavior would occur if the UAV concludes that it {\it must} navigate through a red zone to avoid a crash. Adaptive agents represent a first step for reasoning about such situations in the context of intelligent rebellion/disobedience, and future work can address various forms of constraints, such as competing goals, directives provided by different stakeholders, and ethical prohibitions.

In our current framework, we make no assumptions about the internal representation of directives. We simply require that each directive $\delta \in D$ can be evaluated as a Boolean function on a given state $s$, returning whether the directive is satisfied or violated. In this sense, directives are treated as inputs to our R-HTN agent rather than objects that the agent reasons about. Another avenue for future work is to explore more expressive representations of directives. For example, logic-based formalisms or norm-based representations could allow the agent to infer higher-level properties, such as ethical constraints or normative priorities, and to deliberate about trade-offs among competing directives. Such approaches would also enable the study of directive hierarchies and conflicts, thereby connecting our work more directly with theories of moral reasoning and normative decision-making.

Finally, \cite{mirsky2025artificial} propose three key elements for designing agents that disobey their users, which are also relevant for our Rebel agents. First, agents should be \emph{transparent}, providing explanations for why they are disobeying a command. Second, they should be \emph{justifiable}, meaning that the rationale for disobedience is clear and grounded in defensible principles. Third, they should be \emph{controllable}, allowing the user to override the agent’s disobedience if desired. A promising direction for future work is to endow Rebel agents with these three elements, thereby enhancing their usability and trustworthiness in human-agent interaction.

 


{\parindent -10pt\leftskip 10pt\noindent
\bibliographystyle{cogsysapa}
\bibliography{format}

@inproceedings{coman2014motivation,
  title={Motivation discrepancies for rebel agents: Towards a framework for case-based goal-driven autonomy for character believability},
  author={Coman, Alexandra and Mu{\~n}oz-Avila, H{\'e}ctor},
  booktitle={Proceedings of the 22nd International Conference on Case-Based Reasoning (ICCBR) Workshop on Case-based Agents},
  year={2014}
}

@book{gygax1978advanced,
  title={Advanced dungeons \& dragons},
  author={Gygax, Gary},
  year={1978},
  publisher={TSR Hobbies}
}

@InProceedings{yu:aiide2013,
author     = {Yu, Hong and Riedl, Mark O.},
title      = {Data-Driven Personalized Drama Management},
booktitle  = {Proceedings of the Ninth AAAI Conference on Artificial Intelligence and Interactive Digital Entertainment},
year       = {2013},
month      = {October},
cvsidenote = {56\%},
}

@inproceedings{molineaux2010goal,
  title={Goal-driven autonomy in a Navy strategy simulation},
  author={Molineaux, Matthew and Klenk, Matthew and Aha, David},
  booktitle={Proceedings of the AAAI Conference on Artificial Intelligence},
  volume={24},
  number={1},
  pages={1548--1554},
  year={2010}
}

@article{coman2018ai,
  title={AI rebel agents},
  author={Coman, Alexandra and Aha, David W},
  journal={AI magazine},
  volume={39},
  number={3},
  pages={16--26},
  year={2018}
}

@article{savarimuthu2011norm,
  title={Norm creation, spreading and emergence: A survey of simulation models of norms in multi-agent systems},
  author={Savarimuthu, Bastin Tony Roy and Cranefield, Stephen},
  journal={Multiagent and Grid Systems},
  volume={7},
  number={1},
  pages={21--54},
  year={2011},
  publisher={IOS Press}
}

@inproceedings{sergot2007action,
  title={Action and agency in norm-governed multi-agent systems},
  author={Sergot, Marek},
  booktitle={International Workshop on Engineering Societies in the Agents World},
  pages={1--54},
  year={2007},
  organization={Springer}
}

@inproceedings{georgeff1999belief,
  title={The belief-desire-intention model of agency},
  author={Georgeff, Michael and Pell, Barney and Pollack, Martha and Tambe, Milind and Wooldridge, Michael},
  booktitle={Intelligent Agents V: Agents Theories, Architectures, and Languages: 5th International Workshop, ATAL’98 Paris, France, July 4--7, 1998 Proceedings 5},
  pages={1--10},
  year={1999},
  organization={Springer}
}

@article{huang2025small,
  title={The small-drone revolution is coming—scientists need to ensure it will be safe},
  author={Huang, Xun},
  journal={Nature},
  volume={637},
  number={8044},
  pages={29--30},
  year={2025},
  publisher={Nature Publishing Group UK London}
}

@mastersthesis{Donaldso2014,
    author      = {M.J. Donaldson},
    title       = {Modeling dynamic tactical behaviors in COMBATXXI using hierarchical task networks},
    type        = {MS Thesis}, 
    school = { Naval Postgraduate School},
    year        = {2014}
}

@mastersthesis{Verweij2007,
    author      = {Tim Verweij},
    title       = {A hierarchically-layered
 multiplayer bot system
for a first-person shooter},
    type        = {MS Thesis}, 
    school = { Vrije Universiteit of Amsterdam},
    year        = {2007}
}

@article{gartenberg2014situation,
  title={Situation awareness recovery},
  author={Gartenberg, Daniel and Breslow, Leonard and McCurry, J Malcolm and Trafton, J Greg},
  journal={Human factors},
  volume={56},
  number={4},
  pages={710--727},
  year={2014},
  publisher={Sage Publications Sage CA: Los Angeles, CA}
}

@article{smith1998success,
  title={Success in spades: Using AI planning techniques to win the world championship of computer bridge},
  author={Smith, Stephen JJ and Nau, Dana S and Throop, Thomas A and others},
  journal={AAAI/IAAI},
  volume={98},
  pages={1079--1086},
  year={1998}
}

@article{cardoso2017multi,
  title={A multi-agent extension of a hierarchical task network planning formalism},
  author={Cardoso, Rafael Cau{\^e} and Bordini, Rafael Heitor},
  journal={ADCAIJ: Advances in Distributed Computing and Artificial Intelligence Journal},
  year={2017},
  publisher={Ediciones Universidad de Salamanca (Espa{\~n}a)}
}

@inproceedings{musliner2010priority,
  title={Priority-Based Meta-Control within Hierarchical Task Network Planning},
  author={Musliner, David J and Goldman, Robert P},
  booktitle={2010 Fourth IEEE International Conference on Self-Adaptive and Self-Organizing Systems Workshop},
  pages={282--286},
  year={2010},
  organization={IEEE}
}

@article{georgievski2015htn,
  title={HTN planning: Overview, comparison, and beyond},
  author={Georgievski, Ilche and Aiello, Marco},
  journal={Artificial Intelligence},
  volume={222},
  pages={124--156},
  year={2015},
  publisher={Elsevier}
}

@article{ghallab2014actors,
  title={The actors view of automated planning and acting: A position paper},
  author={Ghallab, Malik and Nau, Dana and Traverso, Paolo},
  journal={Artificial Intelligence},
  volume={208},
  pages={1--17},
  year={2014},
  publisher={Elsevier}
}

@article{yuan2022task,
  title={Task Modifiers for HTN Planning and Acting},
  author={Yuan, Weihang and Munoz-Avila, Hector and Gogineni, Venkatsampath Raja and Kondrakunta, Sravya and Cox, Michael and He, Lifang},
  journal={The Ninth Advances in Cognitive Systems (ACS) Conference},
  year={2022}
}

@article{boussemart2008behavioral,
  title={Behavioral recognition and prediction of an operator supervising multiple heterogeneous unmanned vehicles},
  author={Boussemart, Yves and Cummings, ML},
  journal={Humans operating unmanned systems},
  year={2008}
}

@inproceedings{nau1999shop,
  title={SHOP: Simple hierarchical ordered planner},
  author={Nau, Dana and Cao, Yue and Lotem, Amnon and Munoz-Avila, Hector},
  booktitle={Proceedings of the 16th international joint conference on Artificial intelligence-Volume 2},
  pages={968--973},
  year={1999}
}

@book{chaleff2015intelligent,
  title        = {Intelligent Disobedience: Doing Right When What You're Told to Do Is Wrong},
  author       = {Chaleff, Ira},
  year         = {2015},
  publisher    = {Berrett-Koehler Publishers},
  address      = {Oakland, CA}
}

@article{greene2001fmri,
  title        = {An fMRI investigation of emotional engagement in moral judgment},
  author       = {Greene, Joshua D. and Sommerville, R. Brian and Nystrom, Leigh E. and Darley, John M. and Cohen, Jonathan D.},
  journal      = {Science},
  volume       = {293},
  number       = {5537},
  pages        = {2105--2108},
  year         = {2001},
  publisher    = {American Association for the Advancement of Science},
  doi          = {10.1126/science.1062872}
}

@article{mirsky2025artificial,
  title     = {Artificial Intelligent Disobedience: Rethinking the Agency of {AI} Teammates},
  author    = {Mirsky, Reuth and Shapiro, Daniel and Rahwan, Iyad},
  journal   = {AI Magazine},
  year      = {2025},
  volume    = {46},
  number    = {1},
  pages     = {19--30},
  doi       = {10.1002/aaai.70011},
  publisher = {Wiley}
}

}


\end{document}